\documentclass{article}

\usepackage{microtype}
\usepackage{graphicx}
\usepackage{subcaption}
\usepackage{booktabs} 

\usepackage{algorithm}
\usepackage{algpseudocode}
\usepackage{hyperref}
\usepackage{xurl}  
\usepackage{enumitem}
\usepackage[table]{xcolor} 

\usepackage[accepted]{icml2026}

\usepackage{amsmath}
\usepackage{amssymb}
\usepackage{mathtools}
\usepackage{amsthm}
\usepackage{adjustbox}
\usepackage{multirow}
\usepackage{makecell}

\usepackage[capitalize,noabbrev]{cleveref}
\usepackage{comment}
\theoremstyle{plain}

\theoremstyle{definition}

\theoremstyle{remark}

\usepackage[textsize=tiny]{todonotes}

\icmltitlerunning{MoNe: Modular Neural Memory for Efficient Long Context Inference}

\begin{document}

\twocolumn[
  \icmltitle{MoNe: Modular Neural Memory for Efficient Long Context Inference}

  \icmlsetsymbol{equal}{*}

  \begin{icmlauthorlist}
    \icmlauthor{Wonguk Cho}{q}
    \icmlauthor{Kyubyung Chae}{q}
    \icmlauthor{Tribhuvanesh Orekondy}{q}
    \icmlauthor{Sunghyun Park}{q}
    \icmlauthor{Hyoungwoo Park}{q}
    \icmlauthor{Jeongho Kim}{q}
    \icmlauthor{Arash Behboodi}{q}
    \icmlauthor{Kyuwoong Hwang}{q}
    \icmlauthor{Sungrack Yun}{q}
  \end{icmlauthorlist}

  \icmlaffiliation{q}{Qualcomm AI Research. Qualcomm AI Research is an initiative of Qualcomm Technologies, Inc}

  \icmlcorrespondingauthor{Sungrack Yun}{sungrack@qti.qualcomm.com}
  
  \vskip 0.3in
]

\printAffiliationsAndNotice{}
\begin{abstract}
We present MoNe, a lightweight modular neural memory that attaches to any frozen
pretrained Transformer to enable long-context inference without retraining.
MoNe reads context in fixed-size segments via test-time learning of fast-weight neural
memory networks with layer-localized gradient updates; at inference, the memory generates keys and values from the query tokens alone, with no context tokens re-read. This two-phase design decouples inference cost from context length, achieving $O(N)$
preprocessing and $O(1)$ query cost with peak GPU memory that does not grow with $N$.
At 128K tokens, MoNe reduces both compute and peak GPU memory by approximately 80\%
compared to ICL with only 6.4\% parameter overhead. MoNe generalizes to context lengths far beyond the
backbone's native window, achieving strong performance on needle-in-a-haystack
and word extraction benchmarks from RULER, where ICL degrades sharply.
  \vskip -0.3in
\end{abstract}

\section{Introduction}
Long-context processing has become essential to practical AI applications:
personalized assistants that reason over user history, document QA over lengthy
contracts or medical records, and agentic systems that must synthesize
information spanning tens of thousands of tokens. The dominant approach---feeding the full context as a prompt---scales
quadratically in FLOPs ($O(N^2)$) with context length.
On resource-constrained hardware such as mobile devices, this cost becomes
prohibitive; and for small models that are practical on-device, the problem is
compounded: they struggle to reliably extract information from long prompts even
when those prompts fit within the context window, particularly when tasks
demand complex reasoning across many attended tokens.

\begin{figure}[t!]
  \begin{center}
\centerline{\includegraphics[width=\columnwidth]{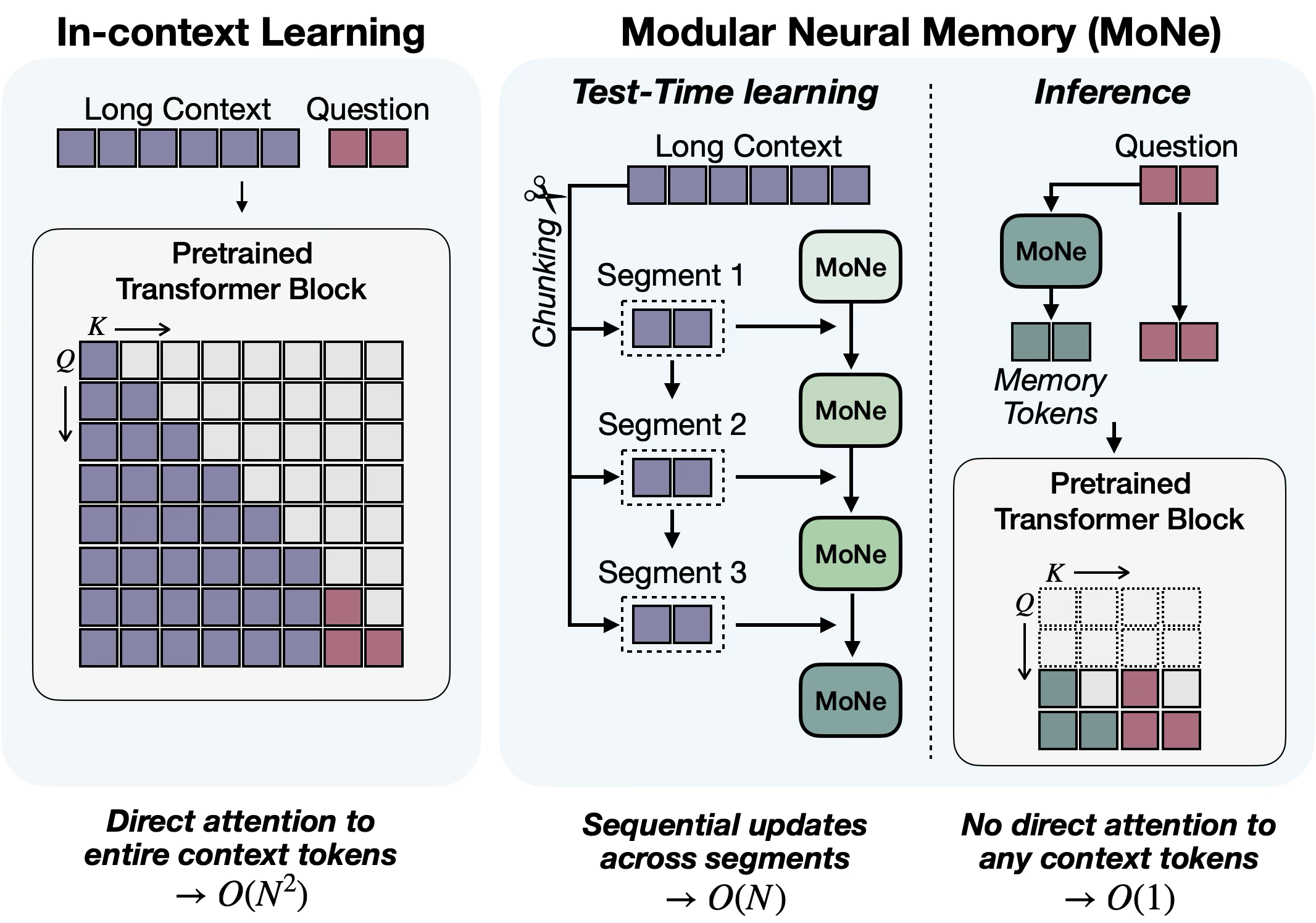}}
\caption{
      Comparison of ICL and MoNe.
      (\textit{Left}) In ICL, question tokens must attend to all context tokens.
      (\textit{Right}) In MoNe, segments are processed sequentially during
      test-time learning to update neural memory; at inference, question tokens
      attend only to memory tokens.
 }
    \label{fig1}
  \end{center}
  \vskip -0.2in
\end{figure}

\begin{figure}[t!]
  \begin{center}
\centerline{\includegraphics[width=1.0\linewidth]{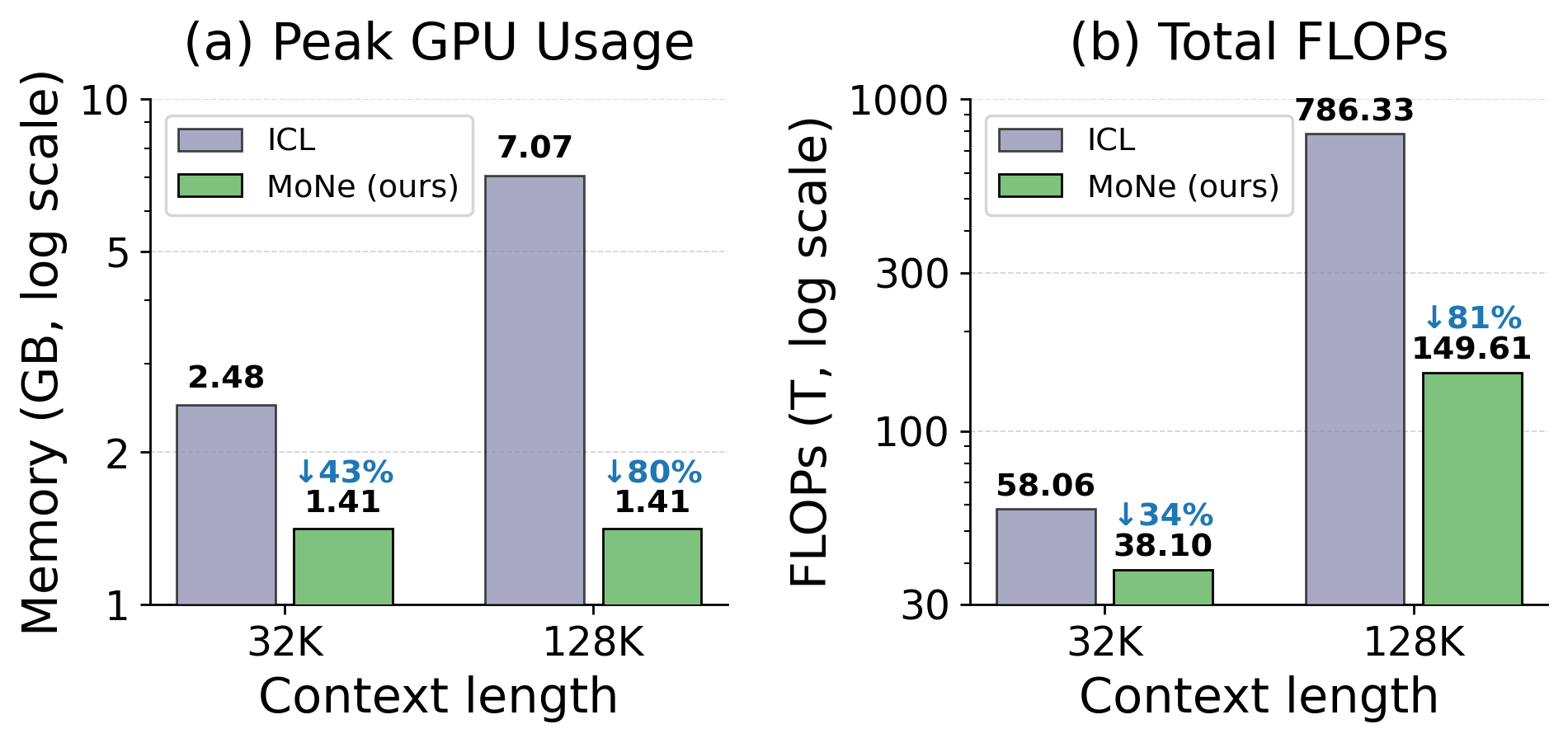}}
    \caption{ End-to-end peak GPU memory and total FLOPs for MoNe (test-time learning + inference) at context lengths of 32K and 128K, in comparison with ICL.}
    \label{fig:cost_analysis}
  \end{center}
  \vskip -0.3in
\end{figure}

As an alternative, Retrieval-Augmented Generation
(RAG)~\citep{lewis2020retrieval,guu2020retrieval} partially addresses the cost
problem by selecting relevant chunks before generation.
However, RAG fundamentally limits reasoning to locally-recoverable facts:
embedding-based retrieval fails when an answer requires integrating multiple
scattered fragments that are individually unremarkable.
On the other hand, recurrent models such as Mamba~\citep{gu2024mamba},
TTT~\citep{sun2025ttt}, and Titans~\citep{behrouz2026titans} achieve
linear-complexity context processing, but they require designing and training
entirely new architectures from scratch and cannot be seamlessly integrated
with existing pretrained Transformers without full retraining.

We propose \textbf{MoNe} (\textbf{Mo}dular \textbf{Ne}ural Memory), a
lightweight plugin module that attaches to any frozen pretrained Transformer
and enables efficient long-context reasoning without retraining the backbone.
As presented in Fig.~\ref{fig1}, MoNe sequentially reads context in fixed-size
segments and performs test-time learning of \emph{fast-weight} neural memory
networks via layer-localized gradient updates---without modifying any backbone weights.
After all context segments are processed, the updated neural memory generates
memory tokens from the question at inference time; these memory tokens serve as keys and values in the frozen backbone's self-attention,
enabling the model to answer from the question alone, with no directly attention to any context tokens.

MoNe's efficiency stems directly from this two-phase design: during test-time learning, context segments of size $T$ are processed sequentially in $O(N)$ total FLOPs; at inference, the query attends only to the memory tokens it generates---not to any context tokens---so inference cost does not scale with $N$, compared to $O(N^2)$ FLOPs for in-context learning (ICL).
The fast weights occupy a constant footprint regardless of context length:
at 128K tokens, this reduces both compute and peak GPU memory by approximately 80\%
compared to ICL, with only 6.4\% parameter overhead, as shown in Fig.~\ref{fig:cost_analysis}.
The final fast-weight state can additionally be reused across multiple queries and
incrementally extended as new context arrives, without reprocessing past content. MoNe generalizes to 128K (a 32$\times$
extrapolation with no additional training) with strong accuracy on S-NIAH, MK-NIAH,
and Frequent Word Extraction benchmarks from RULER, while ICL collapses beyond the
model's native context window.

Our contributions can be summarized as follows:
\begin{itemize}[nosep, leftmargin=2em]
  \item We propose a modular neural memory architecture that plugs into any
    pretrained Transformer without backbone modification, adding only 6.4\%
    parameter overhead.
  \item We introduce a test-time learning procedure with layer-localized
    gradient updates that adapts per-layer fast-weight neural memory networks to encode long contexts in $O(N)$ preprocessing and $O(1)$ query
    cost---compared to $O(N^2)$ for ICL---with natural
    support for incremental context extension and multi-query reuse.
  \item We demonstrate that MoNe trained on contexts up to 4K tokens generalizes
    to 128K (a 32$\times$ extrapolation), achieving strong performance on
    S-NIAH, MK-NIAH, and Frequent Word Extraction where ICL collapses beyond the
    native context window, while reducing total FLOPs and peak GPU memory by
    approximately 80\% at 128K.
\end{itemize}

\section{Related Work}

In this section, we discuss how our work on enabling efficient long-context understanding ties with prior literature.

\paragraph{Long Context Problem.}
Reasoning over long contexts remains a central challenge for LLMs, in part because self-attention scales quadratically with sequence length. 
In practice, long-context settings arise when generation is conditioned on large external context, such as documents in retrieval-augmented generation (RAG)~\citep{vodrahalli2024michelangelo,yen2025helmet}, or on extended reasoning traces, such as those produced during mathematical reasoning~\citep{guo2025deepseek}. 
In this work, we focus on the former setting, where answering a query requires analyzing, understanding, and retrieving relevant information from a long input context. 

Despite increasingly large context windows, LLMs often fail to robustly exploit long contexts, missing facts buried in documents or failing to preserve information from distant positions~\citep{liu-etal-2024-lost,hsieh2024ruler}. 
This limitation is further reflected in the context rot phenomenon~\citep{hong2025context}, where increasing input length degrades a model's ability to use relevant information already present in the context, leading to systematic performance drops even when task-relevant evidence is retained. 
Prior work has sought to improve long-context performance through long-context fine-tuning~\citep{chen2024longlora,gao2025train}, context compression~\citep{yang2025pencil,kontonis2026memento}, and memory-augmented architectures or systems~\citep{chhikara2025mem0,zhou2025mem1,behrouz2026titans}. 
Our work belongs to the memory-augmented line of research, but differs by studying a lightweight neural memory model that is trained at inference time.


\paragraph{Retrieval-Augmented Generation (RAG).}
RAG~\citep{lewis2020retrieval, guu2020retrieval} retrieves relevant documents at inference time
to extend LLMs beyond their context window~\citep{izacard-grave-2021-leveraging, shi2023large}.
In the context of personalization, RAG-based methods select the most relevant fragments of a
user's behavior history to augment the prompt~\citep{salemi-etal-2024-lamp, mysore-etal-2024-pearl},
yielding strong results on personalization benchmarks.
Despite its effectiveness, RAG faces fundamental limitations for personalized on-device agentic
AI.
The computational overhead of repeatedly searching through large local user contexts (e.g.,
lengthy chat histories and app usage logs) at every query is prohibitive on resource-constrained
devices.
More critically, embedding-based retrieval operates on independently chunked segments, making it
ill-suited for long-context reasoning tasks that require identifying and synthesizing multiple
fragmented, interdependent facts distributed across the full user context~\citep{shi2023large, gupta2024rag}.

\paragraph{Test-Time Learning with Neural Memory.}
Test-Time Training (TTT)~\citep{sun2025ttt} has emerged as an alternative paradigm for
long-context modeling, where a small sub-network with rapidly adaptable \textit{fast weights}
is updated at inference time to compress and store past context as neural
memory~\citep{schmidhuber1992fast, behrouz2026titans, yang2024gla, gu2024mamba}.
These fast weights function similarly to recurrent states in RNNs, enabling sub-quadratic
sequence modeling as an alternative to full self-attention.
However, existing TTT methods suffer from low hardware utilization (often below 5\% peak FLOPs)
due to small mini-batch update sizes.
LaCT~\citep{zhang2026lact} addresses this by adopting extremely large chunk updates
(2K--1M tokens), achieving up to 70\% GPU utilization and enabling significantly larger
fast-weight state sizes.
Despite these advances, all such approaches require training new architectures from scratch,
making them difficult to integrate with existing pretrained Transformer-based models.

\section{MoNe: Modular Neural Memory for Pretrained Transformer Attention}
\subsection{Problem Setup}

We consider a pretrained $L$-layer Transformer with hidden dimension $d$ whose
weights are kept completely frozen.
Given a context $\mathbf{C} = (t_1, \ldots, t_N)$ of $N$ tokens and a query
$\mathbf{q}$, our goal is to enable efficient and effective long-context inference
without modifying any backbone parameter.
We partition $\mathbf{C}$ into $S$ non-overlapping segments
$\mathbf{s}_1, \ldots, \mathbf{s}_S$ of $T{=}512$ tokens each, with
$s \in \{1,\ldots,S\}$ and $s{=}0$ denoting the initial state before any
segment is processed.

At each layer $l$, self-attention projects the segment hidden states
$\mathbf{X}^{(l)} \in \mathbb{R}^{T \times d}$ into queries, keys, and values
via frozen weight matrices $\mathbf{W}_q$, $\mathbf{W}_k$, $\mathbf{W}_v$:
\begin{equation}
  \mathbf{Q} = \mathbf{W}_q\mathbf{X}^{(l)}, \quad
  \mathbf{K} = \mathbf{W}_k\mathbf{X}^{(l)}, \quad
  \mathbf{V} = \mathbf{W}_v\mathbf{X}^{(l)},
  \label{eq:attn_proj}
\end{equation}
and computes the attention output as:
\begin{equation}
  \mathrm{Attention}(\mathbf{Q}, \mathbf{K}, \mathbf{V})
  = \mathrm{softmax}\!\left(\frac{\mathbf{Q}\mathbf{K}^\top}{\sqrt{d_h}}\right)\mathbf{V},
  \label{eq:std_attn}
\end{equation}
where $d_h$ is the per-head dimension.
Processing a long context of $N$ tokens incurs $O(N^2)$ FLOPs and requires a KV
cache that grows as $O(N)$, making long-context inference prohibitively expensive
on resource-constrained devices.

\begin{figure}[t!]
  \begin{center}
\centerline{\includegraphics[width=0.9\linewidth]{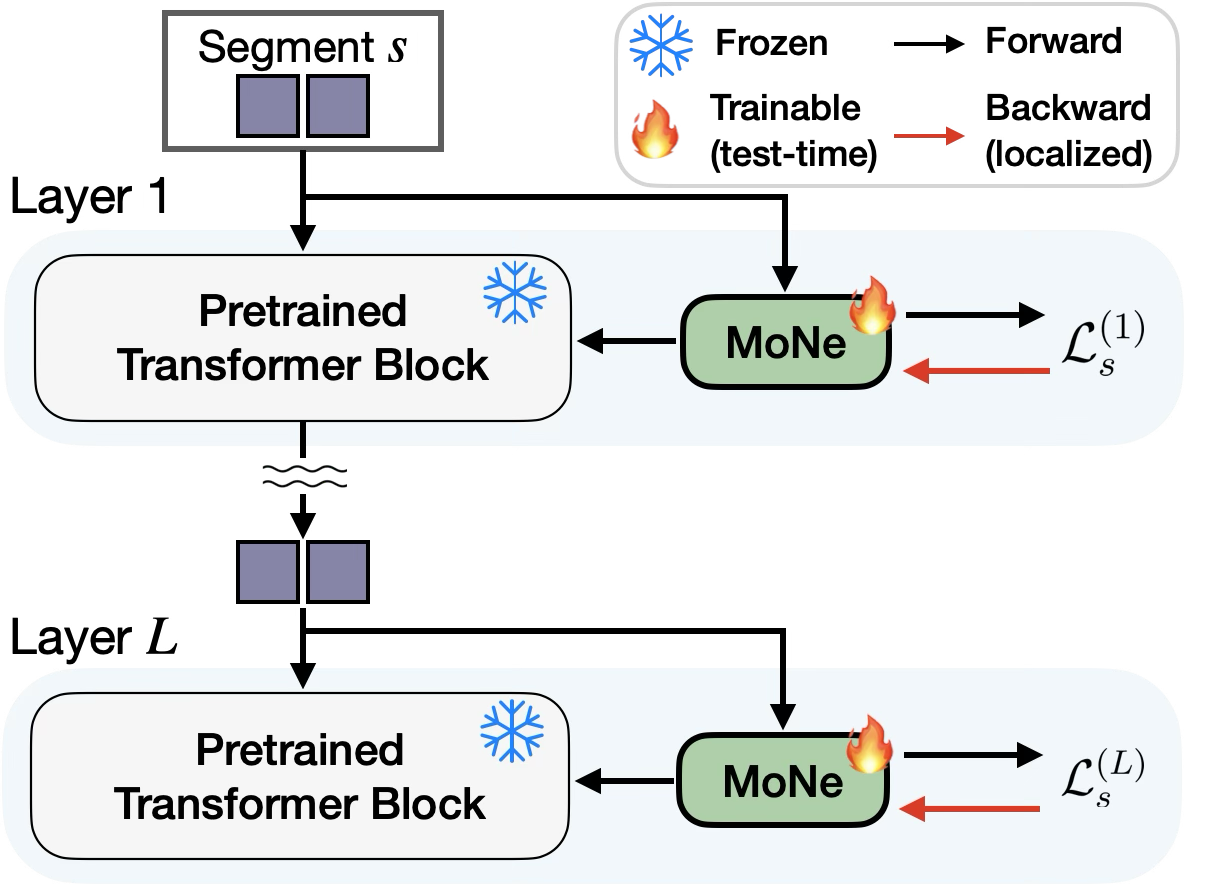}}
    \caption{
An overview of the update process of MoNe during test-time learning. For each layer, neural memory fast weights are updated using the associative memory loss, which is localized and computed entirely from layer $l$'s own forward-pass activations, making the memory update both efficient and truly modular.
    }
    \label{fig2}
  \end{center}
  \vskip -0.2in
\end{figure}

\subsection{Memory Injection into Backbone Attention}
MoNe attaches a fast-weight neural memory to each decoder layer $l$ (Fig.~\ref{fig2}).
Following~\citet{zhang2026lact}, we use a SwiGLU MLP for the memory module of each layer $l$, defined as
$\mathcal{M}(\cdot\,;\,\mathbf{W}^{(l)})$ with fast-weight parameters
$\mathbf{W}^{(l)} = \{\mathbf{W}_{\mathrm{in}}, \mathbf{W}_{\mathrm{gate}}, \mathbf{W}_{\mathrm{out}}\}$:
\begin{equation}
  \mathcal{M}(\mathbf{x};\mathbf{W}^{(l)}) \;=\;
  \mathbf{W}_{\mathrm{out}}^\top\!\Bigl(
    \mathrm{SiLU}\!\bigl(\mathbf{W}_{\mathrm{in}}\,\mathbf{x}\bigr)
    \odot
    \mathbf{W}_{\mathrm{gate}}\,\mathbf{x}
  \Bigr),
  \label{eq:swiglu}
\end{equation}
where all three matrices are $\mathbb{R}^{d_h \times d_h}$ per fast-weight head
and are updated online as context segments are consumed.
Using a nonlinear network rather than a single linear map substantially
increases the expressive capacity of the stored memory, allowing each layer
to encode richer associative structure within the same parameter budget.
At generation time, the query tokens are projected via $\mathbf{W}_q$ and
used to read from the cached fast-weight state $\mathbf{W}^{(l)}_S$
(the state after all $S$ segments have been processed),
producing layer-wise memory tokens:
\begin{equation}
  \mathbf{h}_j = \mathcal{M}\!\left(\mathbf{W}_q\,\mathbf{x}_j;\;\mathbf{W}^{(l)}_S\right).
  \label{eq:read}
\end{equation}
The raw output is normalized per fast-weight head with RMSNorm and scaled by a
learned per-head gate, producing the memory tokens $\mathbf{h}_j$ used below.
These memory tokens are projected into key-value space using the same frozen
backbone matrices $\mathbf{W}_k$ and $\mathbf{W}_v$, and concatenated
alongside the standard attention entries derived from the query tokens:
\begin{equation}
  \mathrm{out} = \mathrm{Attention}\!\left(
    \mathbf{Q},\;
    [\mathbf{K} \,\|\, \mathbf{W}_k\mathbf{h}],\;
    [\mathbf{V} \,\|\, \mathbf{W}_v\mathbf{h}]
  \right),
  \label{eq:attn}
\end{equation}
where $\mathbf{Q}$, $\mathbf{K}$, $\mathbf{V}$ are the standard attention
projections of the query tokens per Eq.~\eqref{eq:attn_proj}.
While $\mathbf{W}_q$, $\mathbf{W}_k$, and $\mathbf{W}_v$ are shared
directly from the backbone without duplication, MoNe adds only small
low-rank adapters~\citep{hu2022lora} to these frozen weights. The memory KV pair has a fixed size of $T$ entries per layer regardless of $N$, keeping the KV-cache footprint constant during both test-time learning and inference.

\subsection{Associative Memory Loss and Update}
\label{sec:mem_update}
During test-time learning, the memory module $\mathcal{M}$ is optimized via an
associative memory loss.
The key-value training targets are produced by $\mathbf{W}_k$ and $\mathbf{W}_v$,
which are frozen backbone weights equipped with meta-trained LoRA adapters:
\begin{equation}
  \mathbf{k}_{s,j}^{(l)} = \mathbf{W}_k\,\mathbf{x}_{s,j}^{(l)}, \quad
  \mathbf{v}_{s,j}^{(l)} = \mathbf{W}_v\,\mathbf{x}_{s,j}^{(l)},
  \label{eq:proj}
\end{equation}
where $j \in \{1,\ldots,T\}$ indexes tokens within segment $s$.
In practice, $\mathbf{k}_{s,j}^{(l)}$ is further passed through SiLU activation
and per-token L2 normalization after projection, and $\mathbf{v}_{s,j}^{(l)}$
through SiLU activation; see Appendix~\ref{app:details} for the full pipeline.
Following~\citet{zhang2026lact}, we define the associative memory loss
as the negative inner product between the memory output and the target value:
\begin{equation}
  \ell_{s,j}^{(l)}\!\left(\mathbf{W}^{(l)}\right) \;=\;
  -\bigl(\mathbf{v}_{s,j}^{(l)}\bigr)^\top\,
  \mathcal{M}\!\left(\mathbf{k}_{s,j}^{(l)};\,\mathbf{W}^{(l)}\right),
  \label{eq:mem_loss}
\end{equation}
with segment-level average $\mathcal{L}_s^{(l)} = \frac{1}{T}\sum_{j=1}^T \ell_{s,j}^{(l)}$.
Minimizing $\ell_{s,j}^{(l)}$ directly imprints the association
$\mathbf{k}_{s,j}^{(l)} \!\to\! \mathbf{v}_{s,j}^{(l)}$ into the fast weights
by maximizing the inner product between the memory output and the target value,
without requiring a prediction error.

The fast weights are updated at each segment via a gradient step:
\begin{equation}
  \mathbf{W}^{(l)}_s \;=\; \mathbf{W}^{(l)}_{s-1} \;-\; \boldsymbol{\mu}_s^{(l)},
  \label{eq:update}
\end{equation}
where $\boldsymbol{\mu}_s^{(l)}$ accumulates gradients of $\ell_{s,j}^{(l)}$ with
per-token learning rates and data-dependent momentum decay (Appendix~\ref{app:details}).

Positional information is incorporated into $\mathbf{k}_{s,j}^{(l)}$ via
segment-local RoPE: each context token is assigned position
$p_j^{\mathrm{local}} {=} j \bmod T$, so position indices always lie in
$[0, T)$ regardless of $N$.
At inference time, query tokens receive positions in $[0, Q)$ where $Q$ is
the query length; since $Q \ll T$, this falls within the same $[0, T)$ range
with no positional interpolation required, allowing MoNe to generalize to
arbitrarily long contexts.
The LoRA adapters and meta-parameters ($\boldsymbol{\eta}^{(l)}$, momentum
projections, output scale) are trained offline via a generation loss on answer tokens
conditioned on the updated memory $\mathbf{W}^{(l)}_S$ and remain frozen during test-time learning and inference, so that
the fast weight update operates as a fixed, pre-learned algorithm at
deployment.

\subsection{Test-Time Learning and Inference}
During \textbf{test-time learning}, segments are consumed sequentially and
the fast weights at each layer $l$ are updated per Eq.~\eqref{eq:update}:
\begin{equation}
  \mathbf{W}^{(l)}_0 \;\to\; \mathbf{W}^{(l)}_1 \;\to\; \cdots \;\to\; \mathbf{W}^{(l)}_S.
\end{equation}
Each update step computes the per-token gradient of $\ell_{s,j}^{(l)}$ locally at
layer $l$:
\begin{equation}
  \nabla_{\mathbf{W}^{(l)}_{s-1}}\ell_{s,j}^{(l)} \;=\;
  -\mathbf{v}_{s,j}^{(l)}\,
  \frac{\partial\,\mathcal{M}\!\left(\mathbf{k}_{s,j}^{(l)};\mathbf{W}^{(l)}_{s-1}\right)^\top}
       {\partial \mathbf{W}^{(l)}_{s-1}},
\end{equation}
carrying no gradient through any other layer:
\begin{equation}
  \frac{\partial \ell_{s,j}^{(l)}}{\partial \mathbf{W}^{(l')}_{s-1}} = \mathbf{0}
  \qquad \forall\, l' \neq l.
\end{equation}
Each update is thus computed entirely from layer $l$'s own forward-pass
activations, making the memory update both efficient and truly modular.
By contrast, updating $\mathbf{W}^{(l)}_{s-1}$ via a standard cross-entropy loss
would require propagating gradients from the output logits back through all
subsequent layers $\mathbf{W}^{(L)}_{s-1}, \ldots, \mathbf{W}^{(l+1)}_{s-1}$
before reaching layer $l$.

During \textbf{inference}, the query tokens are projected via $\mathbf{W}_q$
and the memory is read without any weight update per Eq.~\eqref{eq:read},
producing memory tokens supplied to the frozen backbone in
Eq.~\eqref{eq:attn}.


\begin{table*}[t!]

\caption{%
  Performance of ICL, RAG, and MoNe (ours) on S-NIAH, MK-NIAH,
  and Frequent Word Extraction across varying context lengths.
  Results are grouped by whether the input length falls
  \textit{within} or \textit{beyond} the model's training context
  length (32K tokens), assessing generalization to
  out-of-training-distribution sequence lengths.
}
\label{tab:main_results}
\centering
\tiny
\begin{adjustbox}{width=0.85\textwidth}
\begin{tabular}{clcccccccc}
\toprule
{} & {}
  & \multicolumn{4}{c}{\textit{Within model context length}}
  & \multicolumn{4}{c}{\textit{Beyond model context length}} \\
\cline{3-6} \cline{7-10}
\addlinespace[2pt]
 \bf Task & \bf Method & 4K & 8K & 16K & 32K & 48K & 64K & 96K & 128K \\
\hline
\multirow{3}{*}{S-NIAH}
 & ICL
 & 0.95 & 0.98 & 0.94 & 0.94 & 0.64 & 0.60 & 0.42 & 0.28 \\
 & RAG
 & 0.94 & 0.93 & 0.91 & 0.93 & 0.93 & 0.93 & 0.97 & 0.89 \\
 & \cellcolor{gray!15}\textbf{MoNe (ours)}
 & \cellcolor{gray!15}\textbf{1.00} & \cellcolor{gray!15}\textbf{1.00} & \cellcolor{gray!15} \textbf{0.99} & \cellcolor{gray!15}\textbf{1.00}
 & \cellcolor{gray!15}\textbf{1.00} & \cellcolor{gray!15}\textbf{0.99} & \cellcolor{gray!15}\textbf{0.99} & \cellcolor{gray!15}\textbf{0.96} \\
\hline
\multirow{3}{*}{MK-NIAH}
 & ICL
 & 0.89 & 0.84 & 0.83 & 0.93 & 0.41 & 0.13 & 0.11 & 0.00 \\
 & RAG
 & 0.92 & 0.91 & 0.88 & 0.79 & 0.80 & 0.66 & 0.74 & 0.71 \\
 & \cellcolor{gray!15}\textbf{MoNe (ours)}
 & \cellcolor{gray!15}\textbf{1.00} & \cellcolor{gray!15}\textbf{0.99} & \cellcolor{gray!15}\textbf{0.99} & \cellcolor{gray!15} \textbf{0.99}
 & \cellcolor{gray!15}\textbf{0.99} & \cellcolor{gray!15}\textbf{0.98} & \cellcolor{gray!15}\textbf{0.97} & \cellcolor{gray!15} \textbf{0.94} \\
\hline
\multirow{3}{*}{\shortstack{Frequent\\ Word Extraction}}
 & ICL
 & 0.59 & 0.57 & 0.42 & 0.41 & 0.29 & 0.31 & 0.28 & 0.23 \\
 & RAG
 & 0.58 & 0.61 & 0.60 & 0.61 & 0.61 & 0.59 & 0.60 & 0.60 \\
 & \cellcolor{gray!15}\textbf{MoNe (ours)}
 & \cellcolor{gray!15}\textbf{1.00} & \cellcolor{gray!15}\textbf{1.00} & \cellcolor{gray!15}\textbf{1.00} & \cellcolor{gray!15}\textbf{1.00}
 & \cellcolor{gray!15}\textbf{0.99} & \cellcolor{gray!15}\textbf{0.99} & \cellcolor{gray!15}\textbf{0.96} & \cellcolor{gray!15}\textbf{0.96} \\
\bottomrule
\end{tabular}
\end{adjustbox}
\end{table*}

\section{Experiments}

In this section, we first detail the experimental setup, and follow up by presenting quantitative results and a detailed computation cost analysis.

\subsection{Experimental Setup}

\paragraph{Datasets.}
We evaluate on three tasks drawn from the RULER benchmark~\citep{hsieh2024ruler},
each designed to require precise retrieval or aggregation over the full context
rather than surface-level pattern matching. We use \textbf{S-NIAH} (Single Needle-in-a-Haystack), \textbf{MK-NIAH} (Multi-Key Needle-in-a-Haystack), and \textbf{FWE} (Frequent Word Extraction) tasks.

\paragraph{Evaluation Metric.}
S-NIAH and MK-NIAH use substring exact match (Sub-EM): whether the ground-truth value appears as a substring of the model's output~\cite{asai2024self}. For FWE, we measure variable recall: the fraction of the three target words that appear in the model's output. All tasks are evaluated at context lengths of 4K, 8K, 16K, and 32K tokens
(within the backbone's native window) and 48K, 64K, 96K, and 128K tokens
(beyond the native window). Further training and evaluation details are described in the Appendix~\ref{app:details}.

\paragraph{Baselines.}
We use Qwen2.5-0.5B-Instruct~\citep{qwen2025qwen25technicalreport} as the frozen
backbone for all methods.

\textit{In-Context Learning (ICL)} feeds the full context directly as a prompt,
up to the backbone's native 32K context window; performance degrades sharply beyond
this limit as the model must attend over increasingly many tokens.

\textit{Retrieval-Augmented Generation (RAG)} treats the long context as an external
memory from which relevant information can be retrieved at query time---a natural
baseline that avoids quadratic attention cost by first compressing the context into
a retrievable index~\citep{lewis2020retrieval, guu2020retrieval, izacard-grave-2021-leveraging}.
We split each context into chunks of 128 tokens and use BGE-Large~\citep{chen-etal-2024-m3}
to compute embeddings; the top-$K$ chunks with highest cosine similarity to the query are
concatenated as the prompt.
We evaluate $K \in \{1, 4, 8\}$ and report the best score at each context length.

\subsection{Evaluation Results}

Table~\ref{tab:main_results} reports results of all three methods across three tasks and
eight context lengths, covering both within and beyond the backbone's native 32K window.
The results reveal a consistent pattern: MoNe sustains near-perfect performance across all tasks,
while ICL collapses once the context exceeds the native limit and RAG plateaus on tasks
that require integrating information distributed across the full context.

  \vspace{-5px}

\paragraph{Within the Native Context Window (4K--32K).}
MoNe achieves near-perfect Sub-EM across all tasks and context lengths (0.99--1.00),
substantially outperforming ICL (0.41--0.98) and RAG (0.58--0.93).
ICL's degradation even within the native window is most pronounced on
Frequent Word Extraction, where attending to all tokens is necessary but the
backbone's limited capacity is overwhelmed; MoNe's compressed representation
is more robust.

  \vspace{-5px}

\paragraph{Beyond the Native Context Window (48K--128K).}
ICL degrades sharply once the context exceeds the 32K window: from 0.64 on
S-NIAH at 48K to 0.28 at 128K, and from 0.41 to 0.00 on MK-NIAH.
RAG is more stable but plateaus below 0.97 on S-NIAH and below 0.80 on MK-NIAH,
as embedding-based retrieval struggles with multi-hop facts distributed across
many passages.
MoNe maintains 0.96, 0.94, and 0.96 Sub-EM on S-NIAH, MK-NIAH, and Frequent Word
Extraction at 128K tokens, despite being trained exclusively on contexts up to
4K tokens.
This generalization in context length with no additional training
or positional interpolation is enabled by the segment-local RoPE encoding
(\S\ref{sec:mem_update}), which keeps position indices bounded in $[0,T)$ regardless
of total context length.

\newcommand{\rat}[1]{{\scriptsize(${\downarrow}#1\%$)}}

\begin{table}[t]
\centering
\caption{Computational cost comparison at different context lengths. Values in parentheses denote reduction rate relative to ICL (${\downarrow}$ larger is better).}
\label{tab:cost_compact_ratio}
\small
\begin{adjustbox}{width=\columnwidth,center}
\begin{tabular}{l r l r l}
\toprule
\textbf{Method}
& \multicolumn{2}{c}{\textbf{Peak GPU Usage}}
& \multicolumn{2}{c}{\textbf{Total FLOPs}} \\
\midrule

\multicolumn{5}{c}{\textit{{Context length = 32K}}} \\
\hline
\addlinespace[2pt]
ICL
  & 2.48\,GB & 
  & 58.06\,T & \\
\rowcolor{gray!15} \textbf{MoNe (total)}
  & \textbf{1.41\,GB} & \rat{43}
  & \textbf{38.10\,T} & \rat{34} \\
\hspace{1em}$\llcorner$ Test-time learning
  & 1.41\,GB & 
  & 37.24\,T &  \\
\hspace{1em}$\llcorner$ Inference
  & 1.29\,GB & 
  & 0.86\,T & 
  \\

\midrule
\multicolumn{5}{c}{\textit{{Context length = 128K}}} \\
\hline
\addlinespace[2pt]
ICL
  & 7.07\,GB & 
  & 786.33\,T & \\
\rowcolor{gray!15} \textbf{MoNe (total)}
  & \textbf{1.41\,GB} & \rat{80}
  & \textbf{149.61\,T} & \rat{81} \\
\hspace{1em}$\llcorner$ Test-time learning
  & 1.41\,GB & 
  & 148.97\,T &  \\
\hspace{1em}$\llcorner$ Inference
  & 1.29\,GB & 
  & 0.64\,T & 
  \\
\bottomrule
\end{tabular}
\end{adjustbox}

\vspace{-5px}
\end{table}

\begin{figure*}[t!]
  \begin{center}
\centerline{\includegraphics[width=\textwidth]{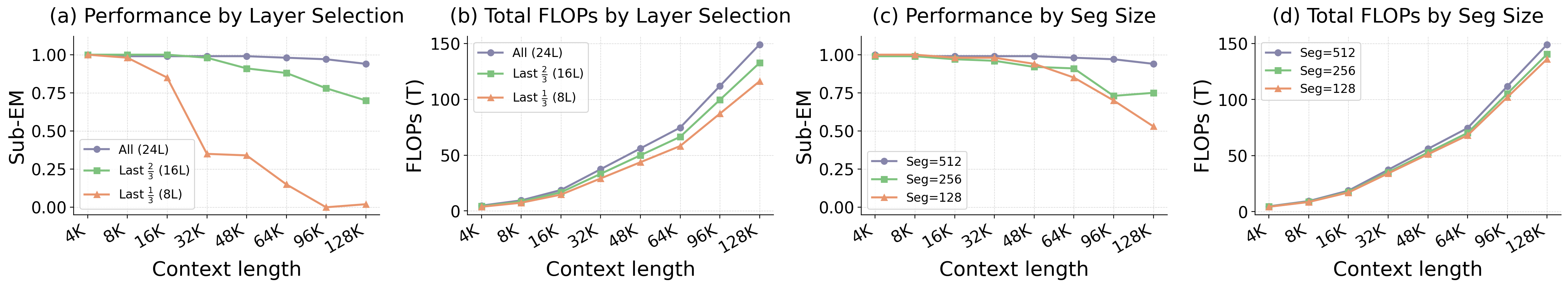}}
    \caption{
Ablation study on performance-efficiency trade-off:     (\textit{Left}) \textbf{Layer selection.}
      Attaching MoNe to all 24 decoder layers achieves the highest accuracy
      across all context lengths; restricting to the last 16 layers
      (last $\tfrac{2}{3}$) yields modest degradation beyond 32K, while
      using only the last 8 layers (last $\tfrac{1}{3}$) collapses beyond 32K. (\textit{Right}) \textbf{Segment size.} $T{=}512$ (default) generalizes from 4K training contexts to 128K evaluation contexts, whereas $T{=}128$ and $T{=}256$ degrade rapidly past 16K, yet offer only marginal efficiency gains over the default.
    }
    \label{fig:sub_em}
  \end{center}

\end{figure*}

\subsection{Computational Cost Analysis}

Table~\ref{tab:cost_compact_ratio} compares the end-to-end computational cost of ICL
and MoNe, including both test-time learning and inference.
At 32K tokens, MoNe reduces peak GPU memory by 43\%
(1.41\,GB vs.\ 2.48\,GB) and total FLOPs by 34\% (38.10\,T vs.\ 58.06\,T).
The advantage becomes more pronounced at longer contexts. At 128K tokens, ICL requires 7.07\,GB and 786.3\,T FLOPs for inference alone, whereas MoNe requires only 1.41\,GB and 149.61\,T FLOPs in total.

Importantly, MoNe’s peak GPU memory is independent of the context length $N$.
During test-time learning, fast-weight updates are computed locally within fixed-size segments
(Eq.~\eqref{eq:update}), resulting in a constant 1.41\,GB.
At inference time, queries attend only to the fixed-size memory tokens,
requiring 1.29\,GB of memory and 0.64\,T FLOPs.

\subsection{Performance-Efficiency Trade-off}

Figure~\ref{fig:sub_em} presents two ablations on MK-NIAH, examining how layer
coverage and segment size each contribute to long-context generalization.

\paragraph{Layer Selection.}
We ablate the effect of attaching MoNe to different subsets of the 24 decoder layers.
Figure~\ref{fig:sub_em}(a) shows that attaching MoNe to all 24 layers achieves Sub-EM
scores of 1.00 / 0.98 / 0.94 at 4K / 64K / 128K. Restricting MoNe to the last 16
layers (the final $\tfrac{2}{3}$) preserves short-context performance (1.00 at 4K)
but degrades at longer contexts (0.88 at 64K, 0.70 at 128K). Restricting further to
the last 8 layers (the final $\tfrac{1}{3}$) leads to a substantial collapse, dropping
to 0.15 at 64K and 0.02 at 128K. 
Each additional group of 8 layers adds only ${\approx}12.6$M parameters
(${\approx}2.1\%$ of backbone size), and Fig.~\ref{fig:sub_em}(b) shows that the
FLOPs overhead is equally modest: at 128K, all-24L incurs 149.1T FLOPs versus 132.6T
for last-16L and 116.1T for last-8L---the restricted configurations reduce FLOPs by
only 11--22\% while suffering a far greater loss in performance.

\paragraph{Segment Size.}
We compare three segment sizes $T \in \{128, 256, 512\}$ across the same context
lengths. Increasing the segment size consistently improves long-context performance:
at 64K / 128K, $T{=}512$ achieves 0.98 / 0.94, outperforming $T{=}256$ (0.91 / 0.75)
and $T{=}128$ (0.85 / 0.53). Although all configurations are trained on contexts of up
to 4K tokens, smaller segments degrade more sharply at longer test contexts.
As shown in Fig.~\ref{fig:sub_em}(d), smaller segments offer only marginal FLOPs
savings; as shown in Fig.~\ref{fig:sub_em}(c), this comes at a substantial cost to
long-context performance. We therefore adopt $T{=}512$, which delivers the strongest
performance among the tested configurations. 

\section{Conclusion}

MoNe augments frozen, pretrained Transformers with long-context capabilities via modular neural memory, requiring no modifications to the backbone weights. Although trained on relatively short sequences, MoNe generalizes to context lengths far exceeding the native window. 
In our experiments, MoNe consistently outperforms ICL, which suffers from sharp performance degradation beyond the training limit. It also surpasses RAG on tasks requiring the synthesis of distributed information. These performance gains are achieved with high efficiency; at 128K tokens, MoNe reduces both FLOPs and peak GPU memory by approximately 80\% compared to ICL. Crucially, its memory footprint remains constant regardless of context length. 
Finally, our ablation studies show that full layer coverage with a segment size of 512 optimizes the performance-efficiency trade-off. MoNe is thus highly effective for resource-constrained applications, such as on-device personalization~\cite{cho2024hollowed,park2026memory} and continual learning~\cite{cho2023complementary, dorovatas2026modular}.

\paragraph{Limitations and Future Work.}
Current experiments validate MoNe's core properties---long-context generalization, constant-memory
inference, and modular integration---using a Qwen2.5-0.5B backbone and the controlled retrieval
tasks in RULER.
Extending these results to larger model scales and to naturalistic long-document workloads such
as single/multi-document QA~\cite{yang2018hotpotqa,rajpurkar2016squad} and real-world conversational histories~\cite{maharana2024evaluating,shaham2023zeroscrolls} is a natural next step: MoNe's
plug-in design requires no architectural changes to accommodate larger backbones, and the
efficiency advantages of constant-memory inference are expected to become even more pronounced
at scale, where KV-cache cost is otherwise extremely prohibitive. A further direction is to apply different LoRA-based approaches~\cite{hu2022lora,hwang2025pica} for each neural memory to improve efficiency and enable flexible management of multiple memory modules.
\bibliography{main}
\bibliographystyle{icml2026}

\newpage
\appendix
\onecolumn
\section{Experimental Details}\label{app:details}

\subsection{Datasets}
From the RULER benchmark~\citep{hsieh2024ruler}, we use the following three tasks:

\begin{itemize}
\item \textbf{S-NIAH} (Single Needle-in-a-Haystack) embeds a single key--value
  pair of adjective-noun compound words at a random depth within a passage of
  filler text; the model must recall the exact value given the key.
  We measure Sub-EM: whether the ground-truth value
  appears as a substring of the model's output.

\item \textbf{MK-NIAH} (Multi-Key Needle-in-a-Haystack) plants four key--value
  pairs of adjective-noun compound words at scattered positions within Paul
  Graham essay excerpts; the model is queried for the value of one specific key
  while the remaining pairs serve as distractors.
  Sub-EM is reported on the queried value.

\item \textbf{Frequent Word Extraction} (FWE) fills the context with synthetic
  6-character words sampled from a Zipfian distribution ($\alpha{=}2.0$), with
  the top-ranked word replaced by a noise token; the model must identify the
  top-3 most frequent non-noise words.
  We measure variable recall: the fraction of the three target words that appear in
  the model's output.
\end{itemize}

With a segment size of 512 tokens, we generate 30K samples for each $N \in
\{2,3,\ldots,8\}$, covering context lengths from 1K to 4K tokens, for offline
training of LoRA adapters and meta-parameters ($\boldsymbol{\eta}^{(l)}$,
momentum projections).
For evaluation, we generate 100 samples per $N \in \{8, 16, \ldots, 256\}$,
ranging from 4K to 128K context lengths.
At test time, context segments are processed sequentially to update the neural
memory; the model then answers using only the final memory state, with no direct
access to the context.

\subsection{Implementation Details}

We use Qwen2.5-0.5B-Instruct as the frozen backbone, attaching a SwiGLU MLP to each decoder layer ($H{=}4$ heads, $d_h{=}224$).
Following \citet{zhang2026lact}, keys and queries undergo affine rescaling, SiLU, and per-token L2 normalization, values SiLU only, and the output is RMSNorm-normalized and SiLU-gated before injection into attention.
Weights are channel-wise L2-renormalized after each update, and the gradient uses a learned per-token learning rate (softplus-activated) and data-dependent momentum decay (sigmoid-activated).
The data-dependent decay $\beta_s^{(l)} \in (0,1)$ is predicted from each chunk's hidden states, allowing the momentum to be selectively reset based on content. For LoRA adapters, rank 128 and $\alpha{=}128$ are used.
We train for one epoch with AdamW ($\beta_1{=}0.9$, $\beta_2{=}0.95$, wd $0.1$), batch 16, lr $10^{-3}$/$5{\times}10^{-5}$ (non-LoRA/LoRA), 200-step warmup and cosine decay to $\eta_{\min}{=}10^{-5}$, gradient clip 1.0, bf16.

\end{document}